\documentclass[10pt,letterpaper]{article}

\usepackage{cogsci}

\cogscifinalcopy 

\usepackage[
    backend=biber,
    style=apa,
    natbib=true,
    doi=false,
    isbn=false,
    url=false,
]{biblatex}
\usepackage{booktabs}
\usepackage{microtype}
\usepackage{pslatex}
\usepackage{float} 
\usepackage{framed}
\usepackage{xcolor}
\usepackage{amsmath}
\usepackage{amssymb}
\usepackage{graphicx}
\usepackage{url}

\definecolor{Blue}{RGB}{0, 0, 255}

\newcommand{\bs}[1]{\boldsymbol{#1}}
\DeclareMathOperator*{\argmax}{arg\,max}

\title{Characterizing tradeoffs between teaching via language and demonstrations in multi-agent systems}
 
\author{Dhara Yu, Noah D. Goodman, Jesse Mu \\ Stanford University \\
        \texttt{\{dharakyu, ngoodman, muj\}@stanford.edu}}

\begin{document}

\maketitle

\begin{abstract}
Humans teach others about the world through \textit{language} and \textit{demonstration}. When might one of these modalities be more effective than the other? In this work, we study the factors that modulate the effectiveness of language vs. demonstration using multi-agent systems to model human communication. Specifically, we train neural network agents to teach via language or demonstration in a grounded communication task, manipulating 1) the inherent difficulty of the task and 2) the competence of the teacher. We find that teaching by demonstration is more effective in the simplest settings, but language is more effective as task difficulty increases, due to its ability to generalize more effectively to unseen scenarios.
Overall, these results provide converging evidence for a tradeoff between language and demonstration as teaching modalities in humans, and make the novel predictions that demonstration may be optimal for easy tasks, while language enables generalization in more challenging settings.

\end{abstract}

\section{Introduction}
One of the defining characteristics of human collective intelligence is the ability for individuals to transmit information about the world to others through the act of teaching \citep{tomasello1999}. While teaching takes many forms, two of the most important modalities are teaching by \emph{language}, i.e.\ describing an underlying rule or desired behavior, or \emph{demonstration}, i.e.\ directly exhibiting the desired behavior. For instance, an experienced chef might instruct a novice by providing a language explanation of a recipe, or demonstrate the process through step-by-step preparation.

When is it optimal to use language vs.\  demonstrations to teach others? Studies of child development suggest that teaching by demonstration may be a cognitively simpler modality: young children tend to teach with demonstration whereas older children and adults teach with language \citep{ellis1982strategies,strauss2002teaching}. On the other hand, language may be better at conveying richer generalizations, especially in tasks requiring more abstract reasoning \citep{tessler2020,SUMERS2023105326, chopra2019}.

While one might conclude from this literature that language is a more effective medium than demonstrations for complex tasks, one confounding factor is human language depends on rich shared conventions and abstractions \citep{clark1996using} that make it especially effective for teaching. Thus, the extent to which these tradeoffs exist for interlocuters learning conventions from scratch is unknown.  Understanding these tradeoffs would not only provide additional evidence for the \emph{a priori} relative effectiveness of teaching modalities, but also inform theories of language evolution grounded in teaching \citep{laland2017origins}: did language become a prominent form of teaching because it was already equipped with useful abstractions? Or could the need to teach others more complex tasks alone motivate language development?
\begin{figure}
    \centering
    \includegraphics[width=\columnwidth]{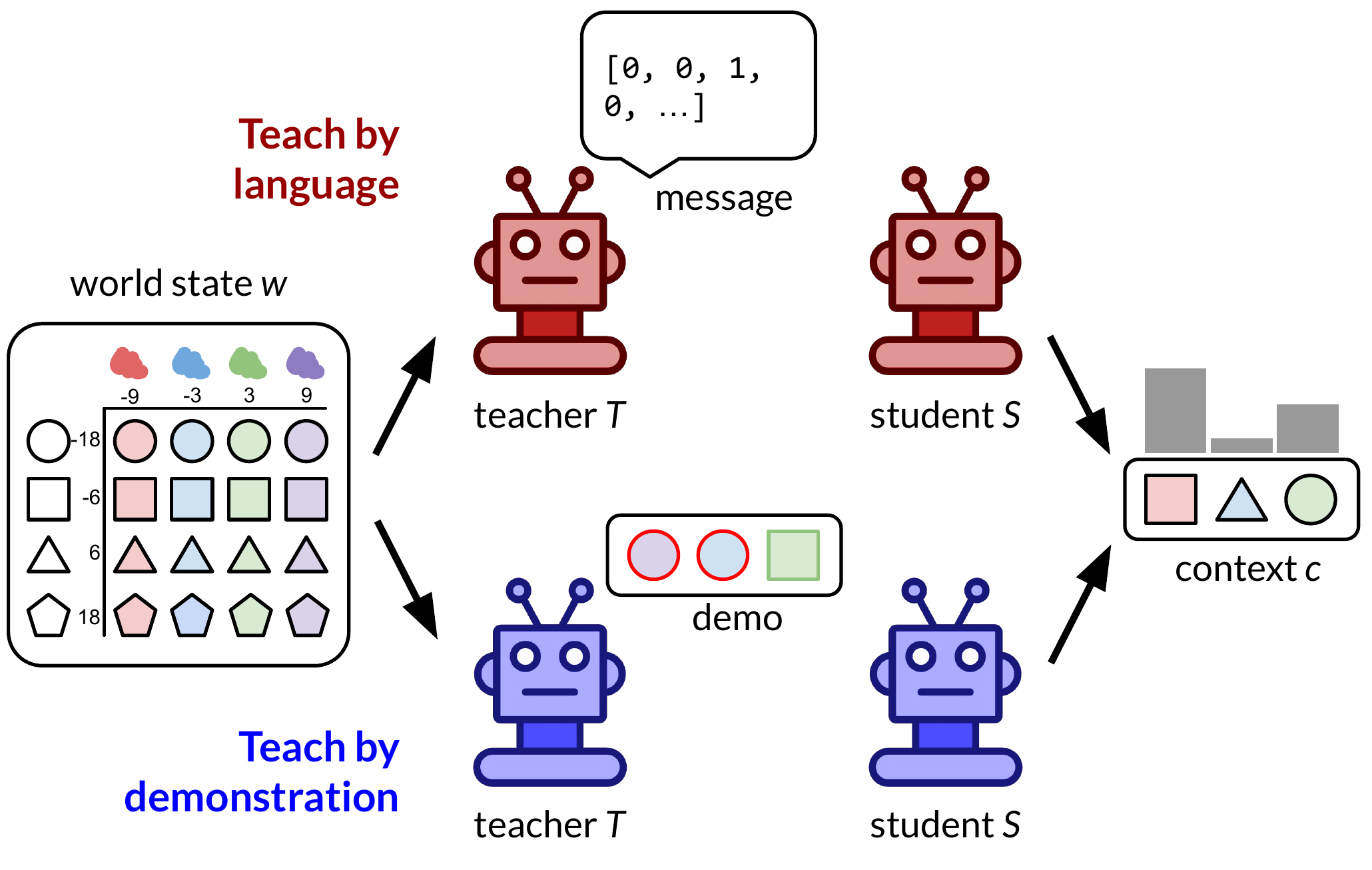}
    \caption{Overview of signaling bandits task in the language (red) and demonstration (blue) conditions. World state $w$ defines the reward associated with each (shape, color) pair. In the language setting, the teacher produces a message represented as one-hot encodings of vocabulary tokens, while in the demonstration setting, the teacher samples an example context and indicates its selection for the context. The student in both conditions selects an item in the context.}
    \label{overview}
\end{figure}

It is difficult to study human teaching in the complete absence of shared conventions: even humans playing abstract and synthetic tasks in the literature \citep{Hawkins2020-HAWCTD-2} quickly learn linguistic conventions that map onto familiar shared concepts. 
Instead, we examine the tradeoffs between language and demonstration using multi-agent communication \citep{lazaridou2020} as a model of human communication. 
Specifically, to study teaching in the absence of conventions, we train neural network agents that learn open-ended communication strategies from scratch to play a game known as the signaling bandits task \citep{sumers2021extending}. Within these simulations, we study how the relative effectiveness of learning language- vs.\ demonstration-based teaching protocols varies based on 1) difficulty of the task and 2) competence of the teacher.



Our experiments show that while agents can devise both language- and demonstration-based teaching protocols, there are clear tradeoffs in the effectiveness of the two mediums. When the task is relatively easy, demonstration agents outperform language agents, but as task difficulty increases, language agents improve in performance, while the ability of demonstration agents drops, corroborating the experimental literature that language is a more useful medium in more challenging settings. Finally, we vary the amount of training data shown to the teacher as a way to tweak teacher competence. The results show that language as as communication medium is more robust to teachers with low competence, with demonstration agents failing to learn in limited-data settings.

Together, our results provide converging evidence consistent with behavioral experiments \citep{chopra2019,SUMERS2023105326} for the increased effectiveness of human language as a teaching modality for more difficult tasks, \emph{even in the absence of conventions}. Our models further predict that i) demonstrations may be optimal for teaching simpler tasks, and ii) language enables more effective transfer of abstractions as teacher knowledge declines. These  simulations support theories that human language may have emerged partially because it is a more effective teaching medium as agents acquire and teach more complex knowledge.

\section{Task}
As a testbed to study the emergence of language- and demonstration-based teaching strategies, we use the signaling bandits task \citep{sumers2021extending} (Figure \ref{overview}). This task generalizes the classic dyadic reference game \citep{Lewis1969-LEWCAP-4} to the multi-armed bandit setting \citep{sutton2018reinforcement}. We pose this as a game to agents in our world, by imagining a \emph{teacher} who knows how to perform the task, and who then tries to teach a \emph{student} how to perform it. Crucially, to succeed at this game the teacher must prepare the student for a variety of situations, unlike in reference games where the teacher only needs to communicate about one specific referring context (but see \citealp{mu2021emergent} for an example of a more generalized reference game). Formally, each game $(\bs{c}, \bs{w})$ is defined by a \emph{context} $\bs{c}$ consisting of a set of objects $\bs{c} = (\bs{c}_1, \dots, \bs{c}_n)$, with each object having a set of boolean features $\phi(\bs{c}_i) \in \{0, 1\}^K$, and a \emph{world state} $\bs{w}$, which describes the reward associated with each feature. For the selected object $\bs{c}_a$ and world state $w$, the reward function $R$ is linear over the object features:
\begin{equation}
R(\bs{c}, \bs{w}, a) = \bs{w}^\intercal \phi (\bs{c}_a).
\end{equation}
Given the true world state $\bs{w}$ and a set of possible utterances $\mathcal{U}$, the teacher $T$'s job is to send an utterance $\bs{u} \in \mathcal{U}$ to the student $S$ such that the student selects the object with the highest utility for some context $\bs{c}$, revealed only to the student. Specifically, given $\bs{\hat{u}} \sim T(\bs{u} \mid \bs{w})$ and $\hat{a} \sim S(\bs{a} \mid \bs{\hat{u}}, \bs{c})$, the reward for both agents is $R(\bs{c}, \bs{w}, \hat{a})$. In this work, we examine how different classes of utterances $\mathcal{U}$, namely language and demonstrations, facilitate success on this task.

\section{Models}
To study the effects of different teaching modalities, we train neural teacher and student agents to play the signaling bandits game. Specifically, the teacher will be trained to both (1) play the game given the ground-truth world state, and (2) send an utterance to the student. In turn, the student will be trained to select the highest utility object given the teacher's message and some game context.


\subsection{Teacher} We can represent the teacher via the conditional distributions:
\begin{align}
    T_{\mathrm{game}}(a \mid \bs{c}, \bs{w}) &\propto \exp (f^T(\bs{w})^\intercal g^T(\bs{c}_a)), \label{eq:tgame} \\
    T_{\mathrm{utt}}(\bs{u} \mid \bs{w}) &= \prod_{i = 1}^{| u |} p_{\mathrm{RNN}}(\bs{u}_i \mid \bs{u}_{< i}\;;\; f(\bs{w})).
\end{align}
Here, $T_{\mathrm{game}}$ is the game-playing portion of the teacher, whose job is to select objects given a context and the known ground-truth world state. Specifically, the probability of a teacher selecting object $\bs{c}_a$ given a context $\bs{c}$ and world state $\bs{w}$ is equal to the dot product between learned representations of the world state $f^T(\bs{w})$ and the corresponding object $g^T(\boldsymbol{c}_a)$, normalized by a softmax decision rule \citep{sutton2018reinforcement}, where $f$ and $g$ are multi-layer perceptrons (MLPs). We train the teacher to play the signaling game itself to learn representations of world states $f^T(\bs{w})$ and contexts $g^T(\bs{c}_a)$ useful for message generation, as well as provide a source of demonstration data for the demonstration agents (introduced later).

Second, $T_{\mathrm{utt}}$ describes the utterance-generating portion of the teacher, where $p_{\mathrm{RNN}}$ is the probability of utterance token $\bs{u}_i$ according to a Gated Recurrent Unit (GRU; \cite{cho-etal-2014-learning}) recurrent neural network (RNN) decoder that conditions on the past messages up to time $i$, $\bs{u}_{< i}$, and the same representation of the world state $f(\bs{w})$ used in $T_{\mathrm{game}}$.

\subsection{Student}
The student is represented similarly to the game-playing portion of the teacher agent, except instead of using the world state $\bs{w}$, it conditions on the teacher's utterance $\bs{u}$: 
\begin{align}
    S(a \mid \bs{u}, \bs{c}) &\propto \exp(f^S(\bs{u})^\intercal g^S(\bs{c}_a)). \label{eq:student}
\end{align}
Again, the probability of object $\bs{c}_a$ is proportional to the dot product between representations of the utterance $f^S(\bs{u})$ and object $g^S(\bs{c}_a)$, similar to $f^T$ and $g^T$ in Equation~\ref{eq:tgame}.

\subsection{Objectives}
Now, given a dataset of games $D = \{ (\boldsymbol{c}_i, \boldsymbol{w}_i) \}_{i = 1}^{M}$, we train our agents jointly via gradient descent to maximize the expected reward $J(S, T)$ of both the teacher and student:
\begin{align}
J_T(c, w) &= \mathbb{E}_{\hat{a}^{T} \sim T_{\mathrm{game}}(a \mid \bs{c}, \bs{w})} \left[ R(\bs{c}, \bs{w}, {\hat{a}^T}) \right], \\
J_{S, T}(c, w) &= \mathbb{E}_{\hat{\bs{u}} \sim T_{\mathrm{utt}}(\bs{u} \mid \bs{w}), \hat{a}^S \sim S(a \mid , \hat{\bs{s}}, \bs{c})}  \left[ R(\bs{c}, \bs{w}, \hat{a}^S) \right], \\
J(S, T) &= \mathbb{E}_{c, w \sim D} \left[ J_T(c, w) + J_{S, T}(c, w) \right].
\end{align}
Here, $J_T$ is the expected reward of the game-playing part of the teacher; $J_{S, T}$ is the expected reward of the student acting on utterances sent from the teacher; and $J(S, T)$ is the combined objective.

To maintain backwards differentiability, we optimize $J(S, T)$ by differentiating through messages sampled from the teacher via the straight-through Gumbel-Softmax trick \citep{Jang2017CategoricalRW, maddison2017}.

\begin{figure}
    \centering
    \includegraphics[scale=0.5]{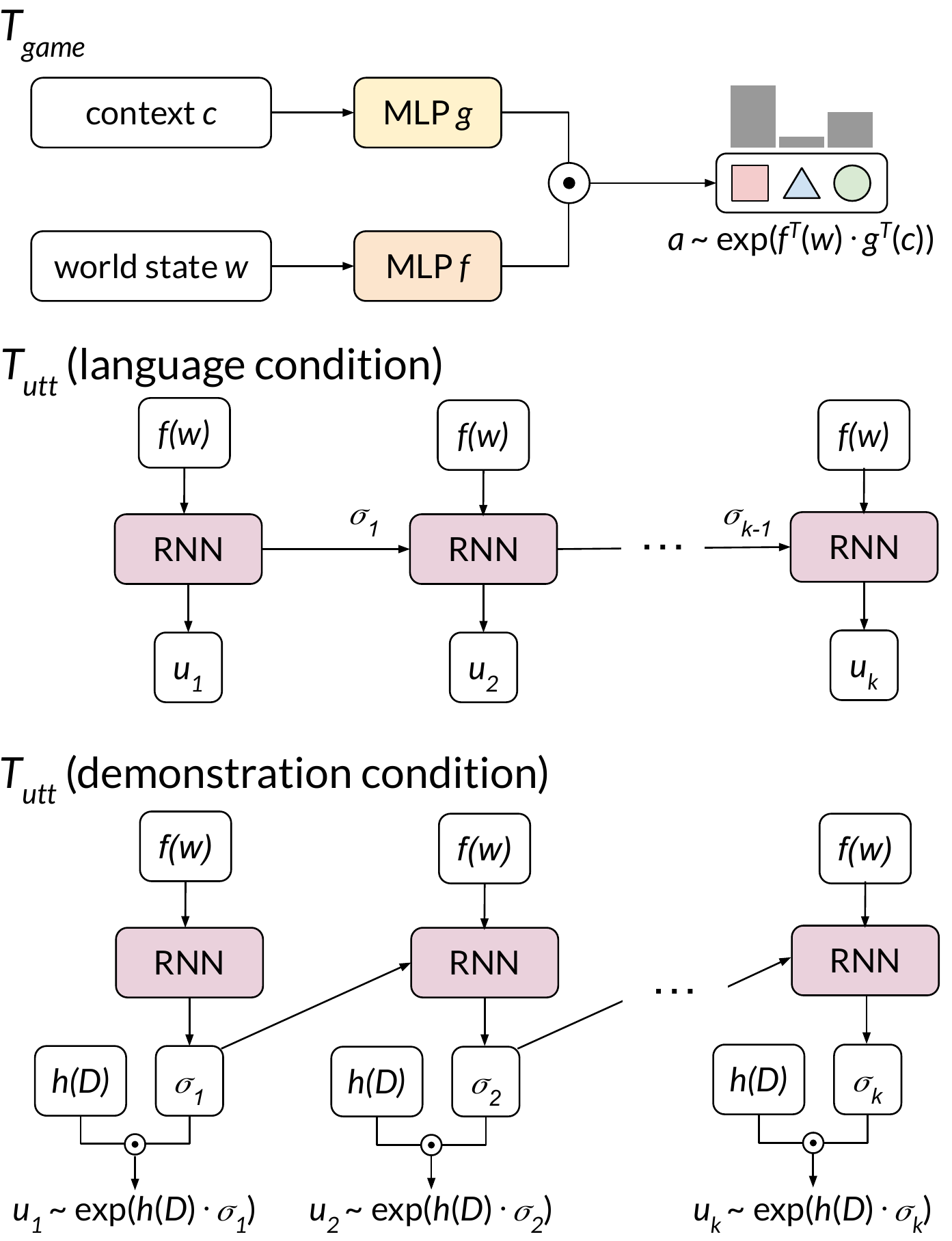}
    \caption{Overview of teacher architecture, including $T_{game}$ (game-playing component) and $T_{utt}$ (utterance-generating component, for both language and demonstrations).}
    \label{architecture}
\end{figure}
\section{Message Conditions}

The differences between agents in our experiments are determined by the space of utterances for communication $\mathcal{U}$, which we describe below.

\subsection{Language}
In the \textbf{language} condition, agents learn a generic discrete communication protocol, as in the emergent communication literature \citep{lazaridou2020}. Here, an utterance $\bs{u}$ is a sequence of up to $K$ discrete tokens: $\bs{u} = (\bs{u}_1, \dots, \bs{u}_k),\, u_i \in \{1, \dots, V\}$, where $k \leq K$ and $V$ is the size of the discrete vocabulary. In this condition, the RNNs used by the teacher and student are standard GRUs as commonly used in natural language processing, with word embedding matrices corresponding to (arbitrary) language tokens, and a language modeling head that projects from the GRU's hidden state at time $i$ to a distribution over tokens for time $i+1$.

\subsection{Demonstrations}
In the \textbf{demonstration} condition, we instead ask the teacher to (1) use $T_{\mathrm{game}}$ to generate a dataset of demonstrations of optimal choices for the given world state, and (2) send a utterance that takes the form of 1 or more demonstrations.

Specifically, for each world state $\bs{w}$, we construct a dataset of demonstrations $D_{\bs{w}} = \{ (\bs{c}^i, \hat{a}^i) \}_{i = 1}^{|\mathcal{C}|}$. For all possible contexts $\bs{c}^i \in \mathcal{C}$, we retrieve the teacher's prediction for this context: $\hat{a}^i = \argmax_a T_{\mathrm{game}}(a \mid \bs{c}^i, \bs{w})$. Then, the teacher sends an utterance to the student where each utterance ``token'' is actually a demonstration: $\bs{u} = (\bs{d}_1, \dots, \bs{d}_k)$, where $\bs{d}_i \in D_{\bs{w}}$.

This means that for demonstration agents, at each time $i$ the teacher RNN selects from a set of demonstrations with the following equation:
\begin{align}
p_{\mathrm{RNN}}(\bs{d}_i \mid \bs{d}_{<i}; f(w)) \propto \exp(h(\bs{d}_i)^\intercal \boldsymbol{o}_i)
\end{align}
i.e.\ a softmax over possible demonstrations in $D_{\bs{w}}$, where $o_i$ is the hidden state of the RNN at time $i$ and $h^T(\bs{d}_i)$ is a learned demonstration encoding (again, a small MLP). Similarly, the student's RNN encoder does not use word embeddings to encode signal tokens $\bs{s}_i$, but rather, a separately learned demonstration MLP
$h^S(\bs{d}_i)$. The demonstration encoder $h^S(\bs{d}_i)$ reuses the context encoder $g^S(\bs{c}_a)$ that is used for the agents' game choice (Equation~\ref{eq:student}).




\section{Experiment Details}

\subsection{Data}
Agents perform the signaling bandits task given a world state with 2 features: color and shape (Figure~\ref{overview}). Both of these two feature can take on $n$ possible values. Each of those $n$ values has an associated reward: the rewards are allocated such that maximum for the a color value is 6 and the minimum for a color value is -6, while the maximum for a shape value is 3 and the minimum is -3, with intermediate values associated with intermediate rewards. For example, a world state with $n=3$ values for each feature (as shown in Figure \ref{overview}A) means that each one of the 3 colors is associated with one of the values in $\{-6, 0, 6\}$, and each one of the 3 shapes is associated with one of the values in $\{-3, 0, 3\}$. For all games, the referring context $c$ over which teachers and students disambiguate consists of 3 objects. Thus for a task setting with $n$ unique values for each of the color and shape features, there are $n! \cdot n!$ unique world states and $\binom{n^2}{3}$ unique possible referring contexts that can be passed to a student as part of a demonstration. 

During training time, teacher agents see world states sampled from 80\% of the possible world states, and are evaluated on world states sampled from the remaining 20\%.  

\subsection{Model and training details}
For the teacher model  (Figure \ref{architecture}), the game-playing portion of the agent is implemented as follows. As aforementioned. $f^{T}$ and $g^{T}$ are MLPs, both of which use ReLU activation functions and produce outputs of 64-d. The utterance-generating portion of the agent differs based on the type of utterance produced. For the language-based teacher, the RNN that defines $p_{RNN}$ produces a 100-d hidden state which is then projected down via a linear layer, over which we sample a vocabulary token using Gumbel-Softmax with $\tau = 1$. For the demonstration-based teacher, $h$ is a multi-layer perceptron that produces a 64-d representation of all possible demonstrations, while the RNN produces a 64-d hidden state representation; we sample a demonstration by taking the dot product of the the demonstration representation and the hidden state via Gumbel-Softmax with $\tau = 1$.

The student model is implemented similarly to the game-playing portion of the teacher: $f^{S}$ is an MLP with output size 64-d, and $g^{S}$ is also an MLP with output size 64-d.

For all simulations, each teacher-student pair is trained with a batch size of 32 games, with 4000 updates for a total of $32 \times 4000 = 128000$ games. The parameters of the teacher and student are jointly updated after each training step, using the Adam optimizer \citep{kingmaba}. For the language experiments, we ran a small search and found empirically that a vocabulary of size $V = 80$ yielded good results, so all language experiments use that value. 

We measure communicative success by computing the normalized reward of the student agent. A reward of 0 indicates random chance performance while a reward of 1 indicates optimal performance (i.e.\ the reward achieved if the student always selected the highest utility object).\footnote{Code for experiments and visualizations is available at \texttt{https://github.com/dharakyu/language-and-demos}.}

\section{Results}

We first present results showing that the language and demonstration agents learn successful communication protocols, in line with expected qualitative patterns. Then, we investigate the tradeoffs between language and demonstration as task difficulty increases and agent competence decreases.

\subsection{Experiment 1: Increasing capacity improves agent performance across modalities}
In our first simulation, we vary the size of the communication channel for language and demonstration agents. For the language agents, this is operationalized through changing the length of the message. For the demonstration agents, this is operationalized through changing the number of demonstrations produced by the teacher and shown to the student. 

\begin{figure}
    \centering
    \includegraphics[scale=0.85]{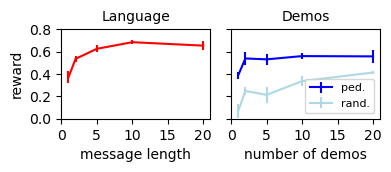}
    \caption{Normalized reward of the student as a function of channel capacity (number of messages for language agents and number of \textit{pedagogical} or \textit{random} demonstrations for demonstration agents) on the validation set. Each point represents the mean of 5 runs. Error bars show standard error of the mean.}
    \label{channel-capacity}
\end{figure}

As a baseline we use \textbf{random demonstration} agents, for which the teacher generates demonstration data, but the demonstrations are randomly selected, so it may not be ideal at conveying key features of the world state. For clarity, we refer to the demonstration agent that selects examples as formulated above as the \textbf{pedagogical demonstration} agent. 

Figure \ref{channel-capacity} shows that as the channel capacity increases for the language and the 2 demonstration modalities, student performance improves. The size of the world state is fixed at $n=4$. For the language student, increasing the length of message leads to gains until message length 10, at which the performance levels off. For the pedagogical demonstration student, performance saturates around $k=2$ demos. For the random demonstration student, performance improves monotonically with increased capacity.

We also observe that the pedagogical demonstration agent outperforms the random demonstration agent. This is consistent with observed behavior in humans, namely that teachers reason about students' knowledge to select examples, rather than choosing at random \citep{Shafto2014ARA, ho2016}.

Finally, Figure 2 shows that the language agent and the pedagogical demonstration agent achieve similar performance with low channel capacity, but as capacity increases, the language agent outperforms the demo agent. This difference in performance cannot be explained by the number of possible signals that can be sent by the language agent vs.\  the demonstration agent: with a message length of 5, the agent can produce $\sum^{k=5}_{i=1} 80^i = \mathcal{O}(10^9)$ unique messages, while a demo agent with $k=5$ can send ${560 \choose 5} = \mathcal{O}(10^{11})$ unique demo combinations,\footnote{This is a conservative lower bound on the number of demos, assuming that agents do not send duplicate demos and that the order in which they are selected doesn't matter.} yet that language agent outperforms the demo agent.

These results confirm that the agents exhibit the same qualitative patterns that we would expect in human participants, namely that increasing the size of the communication channel and selecting teaching examples conditioned on a student both lead to improved performance. Having established this, we can now examine how manipulating task difficulty affects the effectiveness of different teaching modalities.

\subsection{Experiment 2: Effectiveness of teaching by language and demonstration is modulated by task difficulty}

\begin{figure}
    \centering
    \includegraphics[scale=0.85]{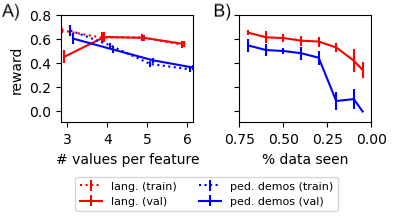}
    \caption{A) Normalized reward on train (dotted line) and validation (solid line) sets as a function of task difficulty. $n$ possible values per feature means that there are $n^2$ unique objects, $n! \cdot n!$ unique reward assignments to those objects, and ${\binom{n^2}{3}}$ unique referring contexts. B) Normalized reward on validation set as a function of the amount of training data. The y-axis is the percent of possible world states seen during training, ranging from 70\% to 5\%. Agents are evaluated on world states sampled from the remaining data.}
    \label{task-difficulty_and-teacher-competence}
\end{figure}

Next, we show how changing task difficulty, by making the task inherently more challenging, affects performance of the language and pedagogical demonstration agents. To operationalize task difficulty, we vary the number of possible values that each feature (color or shape) can take on as defined by the world state, from $n = 3$ to $n = 6$. For example, a world state with $n=4$ has 4 unique color values ($\texttt{red, blue, green, purple}$) and 4 unique shape values ($\texttt{circle, square, triangle, pentagon}$) which each have an associated reward. Intuitively, increasing the number of possible values increases difficulty because the teacher must learn to convey more granular information to maintain the same level of performance. We otherwise keep the size of the communication channel constant: the language-based teacher has a fixed message length of 10, while the pedagogical demonstration-based teacher can send 2 demonstrations.\footnote{These values were chosen based on the saturation points for performance from Experiment 1. We confirmed that the key qualitative patterns hold (namely, that language performance stays relatively flat while demonstration performance declines) even as channel capacity changes by running additional experiments with varying language message/demonstration lengths. The code to run these experiments is available in the publicly-released repository.}

Figure \ref{task-difficulty_and-teacher-competence}A shows that demonstration enables more effective knowledge transmission in the simplest setting $(n = 3)$, but language leads to better performance as task difficulty increases. To show agents' generalization ability, we show the reward earned on both the train and validation sets. In the simplest setting, with a world state containing 3 colors and 3 shapes, the language and pedagogical demonstration agents achieve similar performance on the train set, but the demonstration agent outperforms the language agent on the unseen validation set. The disparity between train and validation for the language agent suggests that the learned communication protocol overfits to the train set with relatively limited data and thus cannot generalize to unseen world states.

Yet as the number of object property values increases, the learning dynamics change. Notably, the language agent no longer exhibits overfitting: train and validation performance is similar. Additionally, the language student achieves similar performance in the $n=4, n=5$ conditions, with a small dropoff for $n=6$. The demonstration student, on the other hand, performs worse as difficulty scales. The reason for this may be that as the number of possible demonstrations becomes combinatorially large as the task is scaled up, whereas the quantity of abstractions that might be conveyed (e.g. ``red is good") increases at a slower rate.

These results are consistent with empirical studies that show teaching by language is more effective for when the task being taught to the student is difficult \citep{SUMERS2023105326}. In addition to matching empirical observations, our simulations make the novel qualitative prediction that teaching by demonstration may be preferred over language for highly simple tasks, in the absence of established linguistic conventions.

\subsection{Experiment 3: Language enables generalization even as teacher competence declines}

In this section, we explore how the competence of the teacher affects student performance across modalities. Teaching is often studied in settings assuming a fully knowledgeable teacher, yet in many naturalistic contexts, teachers do not have oracle knowledge but rather partial, noisy information \citep{Velez2019-VLEIII}.
In our simulations, we operationalize competence by varying the \textit{amount} of data shown to the teacher during training. This is theoretically motivated by the cultural evolution literature, and in particular, the fact that humans see limited data over the course of a lifespan but nonetheless can aggregate this information over generations to develop increasingly complex cultural technologies \citep{tomasello1999}.  

In previous experiments, we fixed the train set to consist of examples sampled from 80\% of all possible world states, while the validation set came from sampling from the remaining 20\%. Here we vary the percentage of world states from which the train set is sampled, to emulate how human teachers have limited access to the structure of a task. Again, we fix the message length for the language agent at 10 and the number of demos for the pedagogical demonstration agent at 2.

As seen in Figure \ref{task-difficulty_and-teacher-competence}B, as the pedagogical demonstration-based teacher sees increasingly limited training data, student performance declines, with a substantial drop occurring at around 0.2. In the lowest-data setting, the student fails to learn at all, exhibiting random chance performance. On the other hand, while the language student's performance declines, it still is able to achieve well above chance performance with highly limited data. These results suggest that language more effectively enables communication about the underlying abstractions that characterize a world state.

\begin{figure}
    \centering
    \includegraphics[scale=0.65]{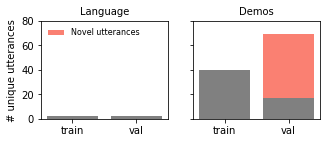}
    \caption{Number of unique utterances (messages or demonstrations) produced across the language and demonstration conditions, on the train and validation sets. Novel utterances are utterances that were produced only on the validation set.}
    \label{num-signals}
\end{figure}

\subsection{Analysis of agent strategies: Language conveys generalizations}
What properties of language enable information transfer even with limited training data? We analyze the messages and demonstrations sent by low-competence teachers that see just 5\% of all possible world states during training, for one run in each condition. Figure \ref{num-signals} shows the \textbf{number of unique utterances}, which are messages for the language agents and demonstrations for the demonstration agents, seen on the train and validation sets, and the \textbf{number of unique novel utterances} (pink portion of validation set bars) produced by the agent on the validation set that were not produced during training. Teachers of both modalities produce far fewer unique signals than unique world states; language-based teachers produce just 2 unique messages in total and no novel ones during test time, while demonstration-based teachers produce substantially more, including several novel ones at test time. 

\begin{figure}
    \centering
    \includegraphics[scale=0.55]{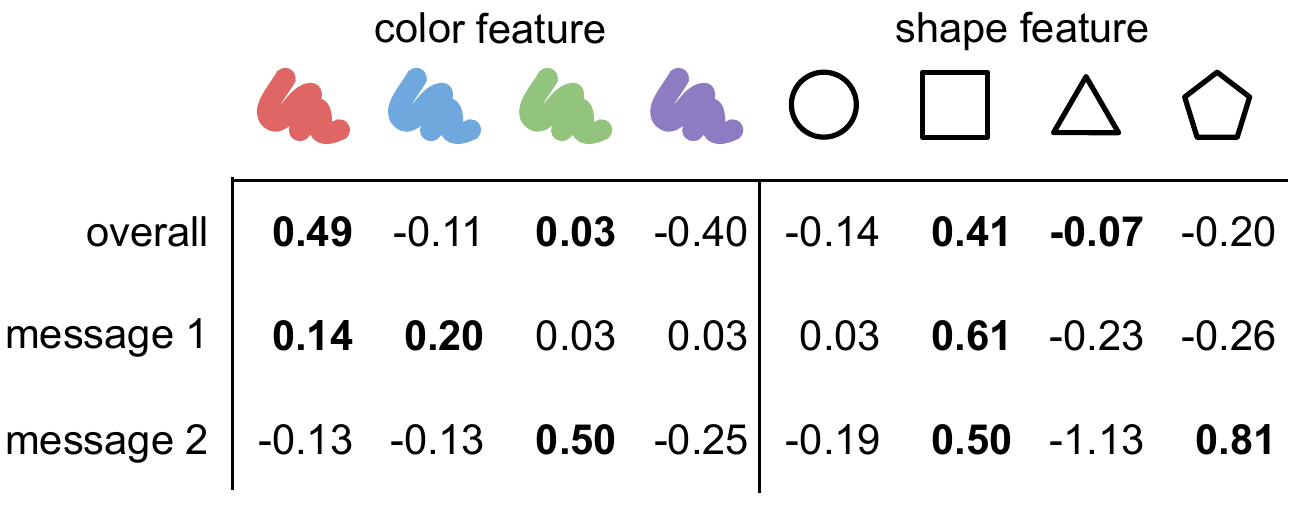}
    \caption{Mean reward of each possible value for each of the color and shape features associated with random message 1, random message 2, and all messages. Within each feature, the two highest rewards are indicated in bold.}
    \label{qualitative-analysis}
\end{figure}

The relatively strong performance of the language agents with so few unique messages (far fewer than the number of world states) necessarily means that the messages themselves contain \textit{abstractions} about the given world state, as opposed to one-to-one mappings between messages and world states. Concretely, the semantics of a message may be closer to ``\texttt{red} and \texttt{blue} are good, \texttt{green} and \texttt{purple} are bad" as opposed to something like ``\texttt{red} has value 6, \texttt{blue} has value 2, \texttt{green} has value -2, \texttt{purple} has value -6". To test this hypothesis, we selected two messages sent by the language-based teacher that is trained on 80\% of the data. Figure \ref{qualitative-analysis} shows the mean reward associated with each of the values for the color (\texttt{red}, \texttt{blue}, etc.) and shape (\texttt{circle}, \texttt{square}, etc.) features, for message 1, message 2, and the full set of messages, produced on the previously-unseen validation set.  For example, the value of 0.14 for \texttt{red} for message 1 indicates that for all reward assignments associated with message 1, the mean value of \texttt{red} was 0.14. We see that while the overall set of validation games is biased toward \texttt{red}, \texttt{green}, \texttt{square} and \texttt{triangle} having relatively high values, message 1 is associated with high rewards for \texttt{red}, \texttt{blue}, \texttt{square} and \texttt{circle}, and message 2 is associated with high rewards for \texttt{green}, \texttt{square} and \texttt{pentagon}. This provides evidence that the teacher is conveying information about the \textit{relative} rewards associated with features, rather than exact quantities, which in of itself is a form of abstraction. This result aligns with findings from behavioral studies that humans use language to communicate abstractions in a teaching setting \citep{SUMERS2023105326, chopra2019}.

\section{Discussion}
We studied the tradeoffs between language- and demonstration-mediated teaching using neural network models of multi-agent interaction. Our results suggest that in the absence of strong communicative conventions, demonstration is more effective for knowledge transmission when the task is simple, while language confers an advantage as task difficulty scales. Furthermore, language maintains the ability to convey generalized knowledge about a world state even when agents have limited access to novel data. A qualitative analysis reveals that language-based agents may accomplish this through communicating higher-level abstractions about the specific world state, such as the relative utilities associated with each feature value.

In addition to providing a formal account that supports findings from empirical studies \citep{SUMERS2023105326}, our models also make novel predictions about the communicative success of teaching with language vs.\  demonstrations in humans. In particular, the modeling predicts that demonstration is more effective and thus may be preferred as a teaching modality for simple tasks. It also predicts that language maintains some degree of effectiveness for teaching on unseen world states even as agents see limited training data, while demonstrations cannot enable generalization to new examples. 

More broadly, these simulations show that under rational pressures, another class of learning agents--neural networks--which are likely optimized in a very different manner compared to humans and in fact have been shown to develop communication systems distinctly unlike human language \citep{chaabouni2019, chaabouni-etal-2020-compositionality}, nonetheless exhibit similar tradeoffs in teaching by language vs.\  demonstration. The existence of these similarities is a form of complementary evidence for empirical results in human studies and bolsters our confidence in using these models as a useful source of predictions. In particular, these results lend credence to the idea that language may have \textit{become} ubiquitous as a teaching modality because of the ease in which it can emerge to convey complex information \citep{laland2017origins}.

In conclusion, this work provides a formal framework and a methodological toolkit for studying the emergence of teaching strategies. Iteratively refining these models and testing their predictions against human behavioral data is an exciting avenue for future work. 

\section{Acknowledgements}
We thank Ted Sumers and the anonymous reviewers for useful feedback.  JM is supported by the Open Philanthropy AI Fellowship. 

\printbibliography

@article{SUMERS2023105326,
title = {Show or tell? Exploring when (and why) teaching with language outperforms demonstration},
journal = {Cognition},
volume = {232},
pages = {105326},
year = {2023},
issn = {0010-0277},
doi = {https://doi.org/10.1016/j.cognition.2022.105326},
url = {https://www.sciencedirect.com/science/article/pii/S0010027722003158},
author = {Theodore R. Sumers and Mark K. Ho and Robert D. Hawkins and Thomas L. Griffiths}
}

@article{tessler2020,
title = {How many observations is one generic worth?}, 
year={2020},
    url = {https://par.nsf.gov/biblio/10159024}, abstractNote = {Generic language (e.g., “Birds fly”) conveys generalizations about categories and is essential for learning beyond our direct experience. The meaning of generic language is notoriously hard to specify, however (e.g., penguins don’t fly). Tessler and Goodman (2019) proposed a model for generics that is mathematically equivalent to Bayesian belief-updating based on a single pedagogical example, suggesting a deep connection be- tween learning from experience and learning from language. Relatedly, Csibra and Shamsudheen (2015) argue that generics are inherently pedagogical, understood by infants as referring to a member of a kind. In two experiments with adults, we quantify the exchange-rate between generics and observations by relating their belief-updating capacity, varying both the number of observations and whether they are presented pedagogically or incidentally. We find generics convey stronger generalizations than single pedagogical observations (Expt. 1), even when the property is explicitly demarcated (Expt. 2). We suggest revisions to the vague quantifier model of generics that would allow it to accommodate this intriguing exchange-rate.}, 
    journal = {Proceedings of the Annual Conference of the Cognitive Science Society}, 
    author = {Tessler, Michael Henry and Bridgers, Sophie and Tenenbaum, Joshua B.}
}

@article{Velez2019-VLEIII,
	journal = {Topics in Cognitive Science},
	year = {2019},
	doi = {10.1111/tops.12388},
	title = {Integrating Incomplete Information with Imperfect Advice},
	pages = {299--315},
	volume = {11},
	number = {2},
	author = {Natalia V\'{e}lez and Hyowon Gweon}
}

@article{ellis1982strategies,
  title={The strategies and efficacy of child versus adult teachers},
  author={Ellis, Shari and Rogoff, Barbara},
  journal={Child development},
  pages={730--735},
  year={1982},
  publisher={JSTOR}
}

@article{strauss2002teaching,
  title={Teaching as a natural cognition and its relations to preschoolers’ developing theory of mind},
  author={Strauss, Sidney and Ziv, Margalit and Stein, Adi},
  journal={Cognitive development},
  volume={17},
  number={3-4},
  pages={1473--1487},
  year={2002},
  publisher={Elsevier}
}

@inproceedings{sumers2021extending,
  author    = {Theodore R. Sumers and
               Robert X. D. Hawkins and
               Mark K. Ho and
               Thomas L. Griffiths},
  title     = {Extending rational models of communication from beliefs to actions},
  booktitle   = {Proceedings of the Cognitive Science Society},

  year      = {2021}
}

@book{Lewis1969-LEWCAP-4,
	title = {Convention: A Philosophical Study},
	publisher = {Cambridge, MA, USA: Wiley-Blackwell},
	year = {1969},
	author = {David Kellogg Lewis}
}

@book{sutton2018reinforcement,
  title={Reinforcement learning: An introduction},
  author={Sutton, Richard S and Barto, Andrew G},
  year={2018}
}

@book{tomasello1999,
    author={Michael Tomasello},
    title={The Cultural Origins of Human Cognition},
    publisher={Harvard University Press},
    year={1999}
}

@misc{lazaridou2020,
  doi = {10.48550/ARXIV.2006.02419},
  
  url = {https://arxiv.org/abs/2006.02419},
  
  author = {Lazaridou, Angeliki and Baroni, Marco},
  
  title = {Emergent Multi-Agent Communication in the Deep Learning Era},
  
  publisher = {arXiv},
  
  year = {2020},
  
  copyright = {arXiv.org perpetual, non-exclusive license}
}

@inproceedings{cho-etal-2014-learning,
    title = "Learning Phrase Representations using {RNN} Encoder{--}Decoder for Statistical Machine Translation",
    author = {Cho, Kyunghyun  and
      van Merri{\"e}nboer, Bart  and
      Gulcehre, Caglar  and
      Bahdanau, Dzmitry  and
      Bougares, Fethi  and
      Schwenk, Holger  and
      Bengio, Yoshua},
    booktitle = "Proceedings of the 2014 Conference on Empirical Methods in Natural Language Processing ({EMNLP})",
    year = "2014",
}

@inproceedings{Jang2017CategoricalRW,
  title={Categorical Reparameterization with Gumbel-Softmax},
  author={Eric Jang and Shixiang Shane Gu and Ben Poole},
  booktitle={5th International Conference on Learning Representations {(ICLR)}},
  year={2017}
}

@inproceedings{maddison2017,
  author    = {Chris J. Maddison and
               Andriy Mnih and
               Yee Whye Teh},
  title     = {The Concrete Distribution: {A} Continuous Relaxation of Discrete Random
               Variables},
  booktitle = {5th International Conference on Learning Representations, {ICLR} 2017},
  year      = {2017},
  bibsource = {dblp computer science bibliography, https://dblp.org}
}

@article{Shafto2014ARA,
  title={A rational account of pedagogical reasoning: Teaching by, and learning from, examples},
  author={Patrick Shafto and Noah D. Goodman and Thomas L. Griffiths},
  journal={Cognitive Psychology},
  year={2014},
  volume={71},
  pages={55-89}
}

@article{Hawkins2020-HAWCTD-2,
	number = {6},
	title = {Characterizing the Dynamics of Learning in Repeated Reference Games},
	doi = {10.1111/cogs.12845},
	volume = {44},
	author = {Robert D. Hawkins and Michael C. Frank and Noah D. Goodman},
	journal = {Cognitive Science},
	year = {2020}
}

@inproceedings{chaabouni-etal-2020-compositionality,
    title = "Compositionality and Generalization In Emergent Languages",
    author = "Chaabouni, Rahma  and
      Kharitonov, Eugene  and
      Bouchacourt, Diane  and
      Dupoux, Emmanuel  and
      Baroni, Marco",
    booktitle = "Proceedings of the 58th Annual Meeting of the Association for Computational Linguistics",
    month = jul,
    year = "2020",
    address = "Online",
    publisher = "Association for Computational Linguistics",
    url = "https://aclanthology.org/2020.acl-main.407",
    doi = "10.18653/v1/2020.acl-main.407",
    pages = "4427--4442",
}

@article{chaabouni2019,
  author    = {Rahma Chaabouni and
               Eugene Kharitonov and
               Emmanuel Dupoux and
               Marco Baroni},
  title     = {Anti-efficient encoding in emergent communication},
  booktitle   = {Neural Information Processing Systems},
  year      = {2019}
}

@inproceedings{ho2016,
 author = {Ho, Mark K and Littman, Michael and MacGlashan, James and Cushman, Fiery and Austerweil, Joseph L},
 booktitle = {Advances in Neural Information Processing Systems},
 editor = {D. Lee and M. Sugiyama and U. Luxburg and I. Guyon and R. Garnett},
 pages = {},
 publisher = {Curran Associates, Inc.},
 title = {Showing versus doing: Teaching by demonstration},
 url = {https://proceedings.neurips.cc/paper_files/paper/2016/file/b5488aeff42889188d03c9895255cecc-Paper.pdf},
 volume = {29},
 year = {2016}
}

@inproceedings{kingmaba,
  author    = {Diederik P. Kingma and
               Jimmy Ba},
  editor    = {Yoshua Bengio and
               Yann LeCun},
  title     = {Adam: {A} Method for Stochastic Optimization},
  booktitle = {3rd International Conference on Learning Representations, {ICLR} 2015,
               San Diego, CA, USA, May 7-9, 2015, Conference Track Proceedings},
  year      = {2015},
  url       = {http://arxiv.org/abs/1412.6980},
  bibsource = {dblp computer science bibliography, https://dblp.org}
}

@inproceedings{chopra2019,
    author={Chopra, Sahil and Tessler, Michael Henry and Goodman, Noah D.},
    title={The first crank of the cultural ratchet: Learning and transmitting concepts through language},
    booktitle={Proceedings of the Cognitive Science Society},
    year={2019}
}

@inproceedings{mu2021emergent,
      title={Emergent Communication of Generalizations}, 
      author={Jesse Mu and Noah Goodman},
      year={2021},
      booktitle={Advances in Neural Information Processing Systems (NeurIPS)}
}

@article{laland2017origins,
  title={The origins of language in teaching},
  author={Laland, Kevin N},
  journal={Psychonomic Bulletin \& Review},
  volume={24},
  pages={225--231},
  year={2017},
  publisher={Springer}
}

@book{clark1996using,
  title={Using language},
  author={Clark, Herbert H},
  year={1996},
  publisher={Cambridge university press}
}

\end{document}